\documentclass[conference]{IEEEtran}
\IEEEoverridecommandlockouts
\usepackage{cite}
\usepackage{amsmath,amssymb,amsfonts}
\usepackage{algorithmic}
\usepackage{graphicx}
\usepackage{textcomp}
\usepackage{xcolor}
\usepackage{geometry}
\usepackage{multirow}
\def\BibTeX{{\rm B\kern-.05em{\sc i\kern-.025em b}\kern-.08em
    T\kern-.1667em\lower.7ex\hbox{E}\kern-.125emX}}
\begin{document}

\title{DMMRL: Disentangled Multi-Modal Representation Learning via Variational Autoencoders for Molecular Property Prediction\\
}
\author{
    \IEEEauthorblockN{Long Xu\textsuperscript{1}, Junping Guo\textsuperscript{1}, Jianbo Zhao\textsuperscript{1},
    Jianbo Lu\textsuperscript{1}, and Yuzhong  Peng\textsuperscript{2*}}
    \IEEEauthorblockA{1. Guangxi Key Lab of Human-machine Interaction and Intelligent Decision, Nanning Normal University, Nanning, China}
    \IEEEauthorblockA{2. College of Big Data and Software Engineering, Zhejiang Wanli University, Ningbo, China}
}

\maketitle
\begin{abstract}
    Molecular property prediction constitutes a cornerstone of drug discovery and materials science, necessitating models capable of disentangling complex structure-property relationships across diverse molecular modalities. Existing approaches frequently exhibit entangled representations—conflating structural, chemical, and functional factors—thereby limiting interpretability and transferability. Furthermore, conventional methods inadequately exploit complementary information from graphs, sequences, and geometries, often relying on naive concatenation that neglects inter-modal dependencies. In this work, we propose DMMRL, which employs variational autoencoders to disentangle molecular representations into shared (structure-relevant) and private (modality-specific) latent spaces, enhancing both interpretability and predictive performance. The proposed variational disentanglement mechanism effectively isolates the most informative features for property prediction, while orthogonality and alignment regularizations promote statistical independence and cross-modal consistency. Additionally, a gated attention fusion module adaptively integrates shared representations, capturing complex inter-modal relationships. Experimental validation across seven benchmark datasets demonstrates DMMRL's superior performance relative to state-of-the-art approaches. The code and data underlying this article are freely available at https://github.com/xulong0826/DMMRL.
\end{abstract}

\section{\textbf{Introduction}} 
Molecular property prediction is vital for drug discovery, enabling efficient identification of promising compounds before synthesis \cite{schneider2020rethinking,walters2020applications,li2022deep,peng2019top}. Accurate predictions of solubility, toxicity, and bioactivity accelerate development while reducing costs. With expanding chemical databases and computational resources, data-driven methods have become essential for understanding structure-property relationships \cite{yang2021deep}. As the chemical space continues to grow exponentially, computational models are increasingly relied upon to prioritize candidates, reduce experimental workload, and guide rational molecular design. These models not only facilitate virtual screening and lead optimization but also enable the exploration of novel chemical scaffolds that may be inaccessible through traditional experimental approaches. In addition, predictive modeling empowers researchers to systematically explore vast chemical spaces, identify structure-activity relationships, and optimize molecular properties with unprecedented efficiency, thereby accelerating the pace of innovation in both pharmaceutical and materials science domains.

Despite these advances, current approaches face two key limitations that hinder their practical utility and scientific insight \cite{xia2023understanding}. First, representation entanglement occurs when structural, chemical, and functional factors become intertwined within learned representations, obscuring which molecular aspects drive specific properties. This entanglement fundamentally constrains model interpretability by conflating distinct causal factors that influence molecular behavior. When multiple structural and electronic properties are encoded in overlapping embedding dimensions, the resulting representations cannot effectively isolate the independent contributions of specific molecular substructures, functional groups, or electronic properties. This not only obscures the mechanistic basis of predictions but also restricts transferability to new tasks where different molecular factors may become relevant. The inability to attribute predictions to specific molecular features ultimately limits the trustworthiness and scientific insight provided by the models, particularly in safety-critical domains like drug discovery where mechanistic understanding is essential for rational design and regulatory approval.

Second, inadequate multi-modal integration persists despite the complementary nature of different molecular representations \cite{wu2023molecular, wu2023improved}. Graph structures capture topological connectivity but lack spatial information; sequence notations encode chemical composition but overlook geometric relationships; conformational representations provide spatial context but may not efficiently encode functional group patterns. Most existing methods either focus exclusively on single modalities, discarding complementary information sources, or employ simplistic concatenation strategies that fail to capture complex inter-modal dependencies and conditional relationships. This deficiency is particularly problematic for properties that emerge from the interplay of multiple structural factors—for instance, molecular binding affinity depends simultaneously on spatial conformation, electronic charge distribution, and functional group arrangement. Effective integration must adaptively weight different modalities according to their task-specific relevance while preserving their complementary nature.

Previous methods exhibit specific shortcomings in addressing these challenges. Traditional descriptor-based approaches offer interpretability but perform poorly with complex patterns and require extensive domain expertise \cite{transfomer, RNN}. These handcrafted features may not generalize well to novel chemical spaces or capture subtle structure-property relationships, limiting their applicability in modern drug discovery pipelines. In contrast, graph neural networks encode molecular topology effectively but often produce entangled representations that obscure the contributions of individual factors \cite{li2021trimnet, cai2022fp, n-Gram}. While GNNs have advanced the field by modeling molecular connectivity, they frequently lack mechanisms to separate property-relevant features from modality-specific noise, resulting in black-box models with limited interpretability. Similarly, standard deep learning approaches create compact representations without addressing disentanglement or multi-modal integration, further exacerbating the challenges of interpretability and adaptability to new tasks.

Building on these limitations, recent advances in variational autoencoders (VAEs) have demonstrated that explicit factor separation can significantly enhance interpretability, generalization, and transferability \cite{liao2023sc2mol}. By enforcing structured latent spaces, VAEs can disentangle shared and private factors, allowing models to focus on the most informative features for property prediction while systematically discarding irrelevant or confounding information. This disentanglement not only improves predictive performance but also provides clearer insights into the molecular determinants of specific properties.

Concurrently, multi-modal learning has emerged as a powerful paradigm for leveraging diverse information sources. Graphs encode connectivity, sequences capture chemical syntax, and geometric representations provide spatial context \cite{zhang2024mvmrl, wang2024multi}. Integrating these modalities can yield richer and more robust molecular representations, enabling models to capture complex structure-property relationships that are inaccessible to single-modality approaches. However, significant challenges remain in disentangling molecular graphs and fusing modalities while maintaining interpretability. Effective multi-modal fusion must not only combine information but also preserve the independence and complementarity of each modality, ensuring that the resulting representations are both informative and interpretable for downstream tasks. Achieving this balance is critical for advancing molecular property prediction and unlocking new opportunities in drug discovery and materials science.

To address these fundamental limitations, we propose DMMRL (Disentangled Multi-Modal Representation Learning), a framework integrating variational autoencoders with multi-modal fusion mechanisms. All in all, the main contributions of DMMRL are as follows.
\begin{itemize} 
    \item We develop a feature disentanglement mechanism using VAE modules that separates shared (common structural) from private (modality-specific) factors within each modality.
    \item We design a gated attention fusion mechanism that adaptively combines shared factors from all modalities, capturing complex inter-modal relationships.
    \item Extensive evaluation on seven benchmark datasets demonstrating DMMRL's superior performance compared to state-of-the-art methods. Ablation studies also demonstrated the effectiveness of the proposed techniques.
\end{itemize}
\section{\textbf{Materials and methods}}\label{sec2}
\subsection{\textbf{Problem formulation}}\label{subsec1} 
\begin{figure*}[htbp]
    \centering
    \includegraphics[width=0.8\textwidth]{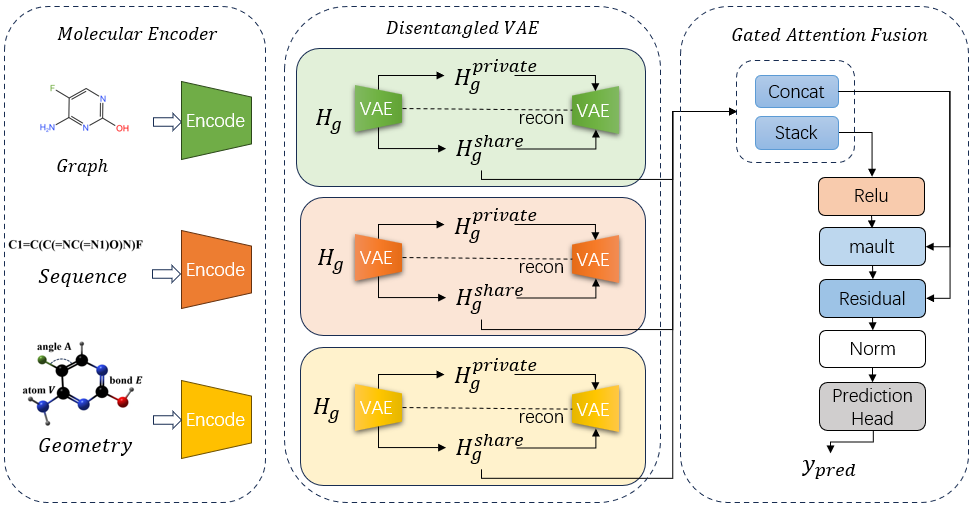}
    \caption{Overview of the DMMRL framework.}
    \label{fig:results}
\end{figure*}

The molecular property prediction problem in a multi-modal context can be formally defined as follows \cite{li2022deep}. Given a collection of molecules $\mathcal{M} = \{m_i\}_{i=1}^N$, each molecule $m_i$ is represented through multiple complementary modalities: a molecular graph $G_i$ that captures atom-bond connectivity patterns and topological structures, a sequence representation $S_i$ (typically SMILES notation) that encodes linear chemical syntax and composition rules, and a geometric conformation $C_i$ that reflects three-dimensional spatial arrangements including bond lengths, angles, and torsional relationships. The objective is to derive a predictive function $f(\cdot)$ that effectively maps these multi-modal inputs to molecular property labels $y_i$, represented as $f(G_i, S_i, C_i) \rightarrow y_i$. The central methodological challenge lies in the effective integration of these complementary information sources while ensuring that the learned representations exhibit both interpretability through factor separation and appropriate disentanglement of property-relevant features from modality-specific artifacts. Figure~\ref{fig:results} illustrates the overall architecture of our proposed DMMRL framework.

\subsection{\textbf{Molecular Encoder Module}}\label{subsec2} 
To process molecular data from different perspectives, we employ three established encoders following methodologies in \cite{wang2024multi}:

\textbf{Sequence Encoder.} For SMILES sequences, a bidirectional LSTM followed by a transformer encoder architecture is implemented. The LSTM effectively captures local chemical contexts from tokenized SMILES strings, extracting sequential dependencies and substructure motifs that are important for chemical interpretation. Subsequently, the transformer encoder leverages self-attention mechanisms to model long-range dependencies and global context, which are essential for comprehending the overall molecular structure and capturing interactions between distant atoms or functional groups within the sequence. This hybrid architecture enables the model to learn both local and global features from SMILES representations, providing a comprehensive encoding of chemical syntax and semantics.

\begin{align}
    h_i^{\text{seq}} &= \operatorname{concat}\left(\overrightarrow{\text{LSTM}}(x_i),\, \overleftarrow{\text{LSTM}}(x_i)\right) \\
    H_s &= \text{Transformer}(\{h_i^{\text{seq}}\}_{i=1}^T)
\end{align}

\textbf{Graph Encoder.} For molecular graphs, a Communicative Message Passing Neural Network (CMPNN) is implemented that systematically updates atom and bond representations through structured information exchange protocols. CMPNN enables each atom and bond to iteratively aggregate information from their neighbors, allowing the encoder to capture complex topological structures, chemical connectivity patterns, and local environments. By facilitating communication between nodes and edges, CMPNN effectively models the intricate relationships present in molecular graphs, such as aromaticity, ring structures, and branching, which are critical for accurate property prediction.

\begin{align}
    m_v^{(k)} &= \text{AGGREGATE}(\{h_{uv}^{(k-1)}: u\in\mathcal{N}(v)\}) \\
    h_v^{(k)} &= \text{COMMUNICATE}(h_v^{(k-1)}, m_v^{(k)})
\end{align}

\textbf{Geometry Encoder.} For three-dimensional conformations, a geometric graph neural network is utilized that incorporates spatial information including distances and angles between atoms. This encoder not only considers the connectivity of atoms but also explicitly models geometric features such as bond lengths, bond angles, and torsional angles, which are crucial for understanding molecular shape and spatial arrangement. By integrating these geometric descriptors, the encoder captures essential three-dimensional arrangement information that complements the topological and chemical features derived from other modalities, enabling the model to better predict properties that are sensitive to molecular conformation.

\begin{align}
    m_v^{(k)} &= \text{AGG}\left(\{h_u^{(k-1)}, h_{uv}^{(k-1)}: u\in\mathcal{N}(v)\}\right) \\
    h_v^{(k)} &= \text{COMBINE}(h_v^{(k-1)}, m_v^{(k)})
\end{align}

\subsection{\textbf{Disentangled Representation VAE Module}}\label{subsec3}
With these specialized encoders extracting modality-specific representations, we introduce a novel disentangled representation learning module based on variational autoencoders (VAEs) to address the fundamental challenge of representation entanglement. Unlike conventional autoencoders or standard disentanglement frameworks—which typically separate latent factors without explicit consideration of modality relevance or predictive utility—our approach systematically decomposes each modality's features into two complementary latent spaces: shared (structure-relevant) and private (modality-specific). This design is grounded in the insight that molecular information comprises both generalizable, property-relevant structural patterns and modality-specific artifacts that may introduce noise.

The key innovation of our method lies in the principled information bottleneck and targeted regularization strategy \cite{hu2024survey}. Rather than simply reconstructing input data or enforcing generic factor separation, our VAE module constrains latent capacity and imposes statistical independence between shared and private spaces, ensuring that only the most predictive molecular features are encoded in the shared space. Orthogonality and alignment losses further enforce that shared representations capture cross-modal consistency while private spaces retain modality-specific details. This is in contrast to ordinary AE-based approaches, which often fail to disentangle predictive and non-predictive factors or allow leakage of irrelevant information into the prediction pathway.

Specifically, for each modality $m \in \{s, g, h\}$ (corresponding to sequence, graph, and geometry representations), the encoder transforms input representations $H_m$ through a pathway that includes normalization, non-linearity, and dimensionality reduction:
\begin{align}
    h_m &= \text{Dropout}(\text{LayerNorm}(\text{ReLU}(\text{Linear}(H_m))))
\end{align}

The encoder then generates parameters (mean and log-variance) for the shared latent distribution, with the private space following an analogous process:
\begin{align}
    \mu_m^{shared}, \log\sigma_m^{shared} &= \text{Linear}_{shared}(h_m)
\end{align}

Rather than deterministically mapping to fixed points, the reparameterization technique facilitates stochastic sampling from the parameterized distribution. This probabilistic encoding enhances regularization effectiveness and prevents posterior collapse by maintaining gradient flow during backpropagation:
\begin{align}
    z_m^{shared} &= \mu_m^{shared} + \sigma_m^{shared} \odot \epsilon_{shared}
\end{align}
where $\sigma_m = \exp(0.5 \cdot \log\sigma_m)$ and $\epsilon \sim \mathcal{N}(0, I)$ represents Gaussian noise. Log-variances are constrained to $[-10, 10]$ for numerical stability while preserving expressivity.

The resulting complementary latent spaces serve distinct functional purposes within the framework:
\begin{itemize}
    \item \textbf{Shared latent space} $z_m^{shared} \in \mathbb{R}^{d_s}$: Encapsulates structural information that remains consistent across modalities and is directly relevant for property prediction. This space captures fundamental molecular characteristics that determine physicochemical properties regardless of representation format, such as electron distribution patterns, stereochemical configurations, and functional group arrangements.
    \item \textbf{Private latent space} $z_m^{private} \in \mathbb{R}^{d_p}$: Encodes modality-specific details that are not directly pertinent to the target property, including representation artifacts, encoding biases, and modality-dependent contextual information. For example, the graph modality's private space may encode specific node ordering schemes, while the sequence modality's private space might capture SMILES syntax conventions that lack direct relevance to physical properties.
\end{itemize}

Moreover, only the shared latent representations are propagated to the downstream prediction head, establishing a strict information bottleneck that compels the model to discard irrelevant modality-specific details. This mechanism not only enhances interpretability and robustness but also ensures that the learned representations are optimally aligned with the property prediction task. By integrating these innovations, our framework achieves superior disentanglement and predictive performance compared to traditional autoencoder-based and factor separation methods.

\subsection{\textbf{Gated Attention Fusion}}\label{subsec4}
Building upon the disentangled representations obtained from the VAE module, we introduce a gated attention fusion mechanism that represents a significant innovation over conventional multi-modal fusion strategies \cite{dey2017gate, niu2021review}. Unlike fixed-weight concatenation or averaging, which assign static and often suboptimal importance to each modality, our approach employs a dynamic, context-aware gating system. This mechanism adaptively modulates the flow of information from each modality based on the specific molecular context and structural characteristics, allowing the model to learn optimal modality contributions for each prediction task.

The core innovation lies in the use of learnable attention gates, which are generated through a multi-layer perceptron (MLP) and normalized via softmax \cite{weerakody2021review}. This enables the model to capture complex inter-modal dependencies and assign probabilistic weights to each modality, ensuring that the sum of contributions equals one. As a result, the fusion process is highly flexible and molecule-specific, reflecting the fact that different molecular properties may be best predicted by different types of information.

The computational pathway of this mechanism consists of three critical stages, each corresponding to the following equations:

\textbf{Context-aware weight generation:}
\begin{align}
    \text{gate\_weights} &= \text{Softmax}(\text{MLP}(\{z_m^{shared}\}_{m=1}^M))
\end{align}
The shared latent vectors from all modalities are first passed through a multi-layer perceptron (MLP), which captures inter-modal dependencies and relative information content. The softmax operation normalizes these outputs into probabilistic weights, ensuring that the sum of contributions from all modalities equals one. This enables the model to dynamically adjust the importance of each modality for every molecule, reflecting the fact that different molecular properties may be best predicted by different types of information.

\textbf{Weighted aggregation of shared representations:}
\begin{align}
    \text{fused} &= \sum_{i=1}^M \text{gate\_weights}_{i} \cdot z_i^{shared}
\end{align}
The shared representations are then combined according to the learned gate weights, amplifying the signal from the most relevant modalities while suppressing less informative ones. This selective fusion is crucial for handling cases where, for example, geometric features dominate in predicting binding affinity, while graph features are more informative for solubility or toxicity.

\textbf{Residual connection integration:}
\begin{align}
    \text{output} &= \text{FFN}\left(\text{fused} + \frac{1}{M} \sum_{i=1}^M z_i^{shared}\right)
\end{align}
A residual connection adds the average of all shared representations to the fused output before passing it through a feedforward neural network (FFN). This design ensures that baseline information from all modalities is preserved, preventing the loss of potentially useful features due to overly aggressive gating. The residual pathway also stabilizes training by maintaining gradient flow and mitigating vanishing gradient issues.

This architecture specifically addresses the heterogeneity of molecular data, where certain modalities may contain more relevant information for specific properties or molecular classes than others. For instance, geometric representations may provide superior information for properties dependent on spatial arrangements (such as protein binding), while graph representations might better capture properties related to functional groups and connectivity patterns. By exclusively utilizing shared representations and excluding private representations from the prediction pathway, this mechanism reinforces the fundamental disentanglement principle while enabling adaptive multi-modal information integration.

\subsection{\textbf{Prediction Head Architecture}}\label{subsec5}
Once multi-modal features are integrated through the gated attention fusion mechanism, a prediction head is employed to map these integrated features to molecular property outputs. The prediction head is implemented as a feedforward neural network with the following structure:
\begin{align}
    \hat{y} &= \text{Linear}(\text{ReLU}(\text{Linear}(output)))
\end{align}

For classification tasks, the output undergoes sigmoid activation to generate probability scores, whereas for regression tasks, raw output values are utilized directly. The prediction head architecture remains deliberately straightforward, thereby focusing the model's capacity on learning meaningful disentangled representations rather than complex prediction mappings.

Task-specific loss functions are applied for prediction:
\begin{align}
    \mathcal{L}_{label} = 
    \begin{cases}
        \frac{1}{|\mathcal{V}|} \sum\limits_{i \in \mathcal{V}} \text{BCE}(y_i, \sigma(\hat{y}_i)) & \text{classification} \\
        \frac{1}{|\mathcal{V}|} \sum\limits_{i \in \mathcal{V}} (y_i - \hat{y}_i)^2 & \text{regression}
    \end{cases}
\end{align}
where $\mathcal{V}$ denotes valid samples, addressing the missing labels issue in molecular datasets.

\subsection{\textbf{Training and Loss Function}}\label{subsec6}
To ensure effective learning of both disentangled representations and accurate property predictions, our objective combines property prediction with principled constraints designed to promote interpretable, disentangled representations \cite{lee2021private,xu2021multi}. The total loss function integrates prediction with multiple specialized regularization terms, each targeting a specific aspect of disentanglement and multi-modal consistency:
\begin{align}
    \mathcal{L}_{total} =\ & \mathcal{L}_{label} + \beta\mathcal{L}_{KL}^{shared} + \lambda\mathcal{L}_{MMD}^{private} \notag \\
    & + \gamma\mathcal{L}_{align} + \delta\mathcal{L}_{ortho} + \eta\mathcal{L}_{recon}
\end{align}
where weights ($\beta$, $\lambda$, $\gamma$, $\delta$, $\eta$) are hyperparameters or learnable coefficients that balance the influence of each loss component during optimization.

\textbf{Variational Disentanglement Losses.} Three complementary regularization terms are employed to ensure the latent spaces are both informative and disentangled:
\begin{align}
    \mathcal{L}_{KL}^{shared} &= \frac{1}{M} \sum_{m=1}^{M} \text{KL}(q(z_m^{shared}|H_m) \| p(z_m^{shared}))\\
    \mathcal{L}_{MMD}^{private} &= \frac{1}{M} \sum_{m=1}^{M} \text{MMD}(z_m^{private}, \mathcal{N}(0, I))\\
    \mathcal{L}_{recon} &= \frac{1}{M} \sum_{m=1}^{M} \|H_m - \hat{H}_m\|_2^2
\end{align}

- $\mathcal{L}_{KL}^{shared}$ is the Kullback-Leibler (KL) divergence between the approximate posterior $q(z_m^{shared}|H_m)$ and the prior $p(z_m^{shared})$ (typically standard normal). This term acts as an information bottleneck, regularizing the shared latent space to prevent overfitting and encouraging the model to encode only the most salient, property-relevant features in $z_m^{shared}$.
- $\mathcal{L}_{MMD}^{private}$ is the Maximum Mean Discrepancy (MMD) between the private latent variables $z_m^{private}$ and a standard normal distribution. This encourages the private latent space to match a target distribution, promoting diversity and flexibility in capturing modality-specific variations while discouraging leakage of shared information into private spaces.
- $\mathcal{L}_{recon}$ is the reconstruction loss, measuring the mean squared error between the original encoder output $H_m$ and its reconstruction $\hat{H}_m$ from the latent variables. This ensures that the latent representations retain sufficient information for accurate reconstruction, counterbalancing the compression imposed by the KL and MMD terms.

\textbf{Representation Relationship Losses.} Two additional terms enforce structural constraints on the relationships between latent spaces:
\begin{align}
    \mathcal{L}_{align} &= \frac{1}{N_{pairs}} \sum_{i<j} \text{InfoNCE}(z_i^{shared}, z_j^{shared})\\
    \mathcal{L}_{ortho} &= \frac{1}{M} \sum_{m=1}^{M} |\langle \hat{z}_m^{shared}, \hat{z}_m^{private} \rangle|
\end{align}

- $\mathcal{L}_{align}$ is an alignment loss based on the InfoNCE contrastive objective, computed over all pairs of shared latent representations from different modalities. By maximizing agreement between $z_i^{shared}$ and $z_j^{shared}$, this term encourages the shared spaces to capture consistent, modality-invariant information, thus facilitating effective multi-modal fusion.
- $\mathcal{L}_{ortho}$ is an orthogonality constraint, penalizing the inner product between normalized shared and private representations within each modality. This enforces statistical independence between $z_m^{shared}$ and $z_m^{private}$, ensuring that property-relevant and modality-specific information are disentangled.

Together, these loss components guide the model to learn representations that are not only predictive but also interpretable, disentangled, and robust across modalities. The careful design and integration of these terms are crucial for achieving the dual goals of high predictive accuracy and meaningful latent structure in multi-modal molecular property prediction.

\section{\textbf{Experimental Settings}}\label{sec3}
\subsection{\textbf{Datasets and Parameter Setting}}

\textbf{Dataset.} To rigorously evaluate the DMMRL framework, comprehensive experiments were conducted on seven benchmark datasets from the MoleculeNet collection \cite{wu2018moleculenetbenchmarkmolecularmachine}. These datasets were specifically selected to represent diverse molecular properties and prediction challenges. The experimental corpus spans various chemical domains including pharmaceuticals, environmental compounds, and small drug-like molecules, thereby providing a robust testbed for assessing model generalizability across diverse chemical spaces.

Table~\ref{tab:dataset_summary} summarizes the key statistics of the datasets used in our experiments, including the number of tasks, sample sizes, task types, and evaluation metrics. The selected datasets cover both classification (e.g., BACE, BBBP, ClinTox, Tox21) and regression tasks (e.g., ESOL, FreeSolv, Lipo), ensuring a comprehensive assessment of model performance across different molecular property prediction scenarios. The diversity in dataset size and property type allows us to systematically analyze the robustness and adaptability of DMMRL under various real-world conditions.
\begin{table}[htbp]
\caption{TABLE I. Summary of datasets.}
\small 
\setlength{\tabcolsep}{3pt} 
\begin{center}
\begin{tabular}{l c c l l}
\hline
\textbf{Dataset} & \textbf{Tasks} & \textbf{samples} & \textbf{Type} & \textbf{Metric} \\
\hline
BACE     & 1  & 1513 & Classification & ROC-AUC \\
BBBP     & 1  & 2040 & Classification & ROC-AUC \\
ClinTox  & 2  & 1478 & Classification & ROC-AUC \\
Tox21    & 12 & 7831 & Classification & ROC-AUC \\
\hline
ESOL     & 1  & 1127 & Regression     & RMSE    \\
FreeSolv & 1  & 639  & Regression     & RMSE    \\
Lipo     & 1  & 4200 & Regression     & RMSE    \\
\hline
\end{tabular}
\label{tab:dataset_summary}
\end{center}
\end{table}

\textbf{Parameter Setting.} To ensure methodological consistency and fair comparison with existing approaches, datasets were partitioned using random splitting according to established MoleculeNet protocols, with an 8:1:1 ratio for training, validation, and test sets, respectively. Each experiment was replicated 10 times with different random seeds to ensure statistical robustness, with results reported as mean and standard deviation of area under the receiver operating characteristic curve (ROC-AUC) for classification tasks and root mean squared error (RMSE) for regression tasks.

For DMMRL, we set the number of training epochs to 200 and use a batch size of 64 unless otherwise specified. The Noam learning rate scheduler is adopted, with an initial and final learning rate of $1\times10^{-3}$ or $1\times10^{-4}$, and a maximum learning rate of $2\times10^{-3}$ or $2\times10^{-4}$, selected based on validation performance. All regularization weights, including $\beta$ for the KL divergence term, $\lambda$ for the MMD term, $\gamma$ for the alignment loss, $\delta$ for the orthogonality constraint, and $\eta$ for the reconstruction loss, are learnable parameters with initial values set to 0.1. All coefficients are selected based on validation performance and grid search to ensure optimal model training. All hidden dimensions are set to 256. The dimensions of shared and private latent features are selected according to the specific task and dataset, and these settings can be found in detail in our released code. All activation functions are ReLU, and Layer Normalization is applied throughout the network \cite{chen2020dynamic, ba2016layernormalization}. Residual connections and feedforward neural networks are incorporated for feature enhancement and stabilization \cite{he2016deep}. The Adam optimizer is used for training, and gradient clipping is applied to ensure stability \cite{kingma2014adam}.

All experiments were conducted on a workstation equipped with an Intel Core i7-14700KF processor (20 cores, 28 threads), 64GB of system memory, and an NVIDIA GeForce RTX 4060 Ti GPU with 16GB VRAM. Under these settings, training the full DMMRL model on all datasets required approximately 2 days to complete.

\begin{table*}[htbp]
\caption{Molecular property prediction results on classification and regression datasets with baselines and DMMRL.}
\small 
\setlength{\tabcolsep}{3pt} 
\begin{center}
\begin{tabular}{l|cccc|ccc}
\hline
\multirow{2}{*}{\textbf{Methods}} 
& \multicolumn{4}{c|}{\textit{ROC-AUC $\uparrow$ (Higher is better)}} 
& \multicolumn{3}{c}{\textit{RMSE $\downarrow$ (Lower is better)}}\\
\cline{2-8}
& \textbf{BACE} & \textbf{BBBP} & \textbf{ClinTox} & \textbf{Tox21} 
& \textbf{ESOL} & \textbf{FreeSolv} & \textbf{Lipophilicity} \\
\hline
RNN      & - & 0.902$\pm$0.015 & 0.915$\pm$0.009 & 0.806$\pm$0.007 & 0.743$\pm$0.020 & 1.108$\pm$0.146 & 0.770$\pm$0.025 \\
Transformer & - & 0.944$\pm$0.011 & \underline{0.954$\pm$0.003} & 0.813$\pm$0.013 & 0.767$\pm$0.079 & 1.021$\pm$0.102 & 0.900$\pm$0.023 \\
GCN      & - & 0.690$\pm$0.036 & 0.807$\pm$0.051 & 0.829$\pm$0.041 & 0.970$\pm$0.050 & 1.400$\pm$0.135 & - \\
Weave    & - & 0.671$\pm$0.065 & 0.832$\pm$0.023 & 0.820$\pm$0.024 & 0.610$\pm$0.055 & 1.220$\pm$0.250 & - \\
MPNN     & - & 0.910$\pm$0.032 & 0.881$\pm$0.037 & \underline{0.844$\pm$0.014} & 0.702$\pm$0.042 & 1.242$\pm$0.249 & 0.645$\pm$0.075 \\
N-Gram   & - & 0.912$\pm$0.013 & 0.855$\pm$0.037 & 0.842$\pm$0.027 & 1.100$\pm$0.160 & 2.512$\pm$0.190 & 0.876$\pm$0.033 \\
GROVER   & - & 0.955$\pm$0.003 & 0.929$\pm$0.178 & 0.842$\pm$0.009 & 0.911$\pm$0.116 & 1.987$\pm$0.072 & 0.643$\pm$0.030 \\
TrimNet  & 0.841$\pm$0.043 & 0.889$\pm$0.020 & 0.948$\pm$0.030 & - & 0.770$\pm$0.071 & 1.639$\pm$0.406 & 1.202$\pm$0.032 \\
FP-GNN   & \underline{0.881$\pm$0.028} & 0.935$\pm$0.027 & 0.840$\pm$0.038 & 0.815$\pm$0.024 & 0.675$\pm$0.332 & 0.905$\pm$0.649 & 0.625$\pm$0.152 \\
GraSeq   & - & 0.942$\pm$0.012 & 0.918$\pm$0.005 & 0.810$\pm$0.004 & 0.652$\pm$0.039 & 0.865$\pm$0.032 & 0.648$\pm$0.041 \\
MvMRL         & 0.891$\pm$0.019 & 0.962$\pm$0.011 & \textbf{0.975$\pm$0.019} & \textbf{0.845$\pm$0.011} & 0.601$\pm$0.055 & \underline{0.832$\pm$0.128} & 0.634$\pm$0.018 \\
SGGRL         & 0.917$\pm$0.020 & \underline{0.967$\pm$0.010} & 0.956$\pm$0.016 & 0.837$\pm$0.013 & \underline{0.575$\pm$0.057} & 0.847$\pm$0.116 & \underline{0.617$\pm$0.025} \\
DMMRL(Ours)   & \textbf{0.925$\pm$0.012} & \textbf{0.968$\pm$0.013} & 0.935$\pm$0.021 & 0.842$\pm$0.013 & \textbf{0.535$\pm$0.068} & \textbf{0.825$\pm$0.089} & \textbf{0.599$\pm$0.033} \\
\hline
\end{tabular}
\label{tab:results}
\end{center}
\end{table*}

\subsection{\textbf{Method Comparison}}
For comprehensive evaluation, DMMRL was systematically compared with twelve established baseline approaches spanning sequence-based, graph-based, geometry-based, and multi-modal molecular representation methodologies. Specifically, Transformer \cite{transfomer} and RNN \cite{RNN} represent SMILES sequence-based models. TrimNet \cite{li2021trimnet}, FP-GNN \cite{cai2022fp}, GCN \cite{GCN}, Weave \cite{Wrave}, MPNN \cite{MPNN}, and N-Gram \cite{n-Gram} constitute graph neural network methods of varying architectural complexity. Additionally, GROVER \cite{GROVER} implements sophisticated message-passing operations on directed graphs. GraSeq \cite{guo2020graseq} combines molecular SMILES and graph representations in a dual-modality approach. Finally, MvMRL \cite{zhang2024mvmrl} and SGGRL \cite{wang2024multi} represent state-of-the-art multi-modal approaches that integrate graph and three-dimensional geometry information.

\subsection{\textbf{Results and Analysis}}
Our comprehensive evaluation demonstrates that DMMRL achieved superior performance on five of the seven benchmark datasets (BACE, BBBP, ESOL, FreeSolv, and Lipophilicity), as shown in Table~\ref{tab:results}. For bioactivity classification tasks, DMMRL reached 92.5\% and 96.8\% ROC-AUC on BACE and BBBP respectively, representing improvements of 0.8 and 0.1 percentage points over the next best methods. For physicochemical property regression tasks, RMSE reductions of 6.9\%, 0.8\%, and 2.9\% were observed on ESOL, FreeSolv, and Lipophilicity respectively compared to state-of-the-art alternatives.

Notably, performance advantages were most pronounced on structurally diverse datasets with sparse annotations (BACE, FreeSolv), where effective feature disentanglement becomes particularly critical. On the BACE dataset, which consists of binding affinity data for $\beta$-secretase inhibitors spanning diverse chemical scaffolds, DMMRL demonstrated a substantial 0.8 percentage point improvement in ROC-AUC over the previously best-performing model. Similarly, for the challenging FreeSolv dataset (hydration free energy predictions with only 639 samples), DMMRL reduced RMSE by 0.8\% compared to the next best method. This empirical finding confirms our theoretical proposition that disentanglement mechanisms are especially valuable when training data is limited or exhibits high structural variability, as they enable the model to more effectively identify and isolate generalizable structure-property relationships from modality-specific noise.

The performance improvement pattern across datasets correlates strongly with chemical diversity and data sparsity metrics, suggesting that DMMRL's disentanglement approach specifically addresses the challenges of learning from heterogeneous molecular data. By forcing the model to encode only the most relevant molecular features in the shared latent space, DMMRL effectively implements an inductive bias that favors generalizable patterns over dataset-specific correlations or artifacts.

Interestingly, DMMRL exhibited consistently low standard deviations across multiple evaluation runs, particularly for BBBP (±0.013) and Tox21 (±0.013), indicating that the model produces reliable predictions across different data subsets—a critical advantage for early-stage drug discovery applications where predictive reliability across diverse chemical scaffolds is essential for effective compound prioritization. The combined improvements in both accuracy and consistency demonstrate that disentangled representations enable more robust structure-property modeling, particularly for the complex, multi-factorial molecular properties that characterize pharmaceutical and materials science applications.

\begin{table}[htbp]
\caption{Ablation study of DMMRL components across classification and regression tasks.}
\small 
\setlength{\tabcolsep}{3pt} 
\centering
\begin{tabular}{l|ccc}
\hline
\textbf{Dataset} & \textbf{LBL} & \textbf{BOT} & \textbf{ALL (DMMRL)} \\
\hline
\multicolumn{4}{l}{\textit{Classification Tasks (ROC-AUC $\uparrow$)}} \\
\hline
BACE & 0.906$\pm$0.117 & \underline{0.911$\pm$0.012} & \textbf{0.925$\pm$0.012} \\
BBBP & 0.931$\pm$0.011 & \underline{0.966$\pm$0.048} & \textbf{0.968$\pm$0.013} \\
ClinTox & 0.917$\pm$0.036 & \underline{0.918$\pm$0.018} & \textbf{0.935$\pm$0.021} \\
Tox21 & 0.822$\pm$0.032 & \underline{0.840$\pm$0.022} & \textbf{0.842$\pm$0.013} \\
\hline
\multicolumn{4}{l}{\textit{Regression Tasks (RMSE $\downarrow$)}} \\
\hline
ESOL & 0.614$\pm$0.066 & \underline{0.563$\pm$0.027} & \textbf{0.535$\pm$0.068} \\
FreeSolv & 0.878$\pm$0.175 & \underline{0.855$\pm$0.085} & \textbf{0.825$\pm$0.089} \\
Lipophilicity & 0.615$\pm$0.030 & \underline{0.612$\pm$0.016} & \textbf{0.599$\pm$0.033} \\
\hline
\end{tabular}
\label{tab:ablation_combined}
\end{table}

\subsection{\textbf{Ablation Studies}}
To systematically validate the contribution of each architectural component motivated by our theoretical framework, we conducted detailed ablation experiments with three progressively more complete model configurations: LBL (prediction pathway only, without disentanglement or regularization), BOT (adding variational bottleneck with KL and MMD regularization), and ALL (complete DMMRL with alignment and orthogonality constraints). Table~\ref{tab:ablation_combined} summarizes these results across all datasets.

The ablation analysis revealed consistent and incremental performance improvements with each additional component. The transition from LBL to BOT yielded particularly substantial gains across multiple datasets (3.5 percentage point ROC-AUC improvement for BBBP; 8.3\% RMSE reduction for ESOL), demonstrating that the variational constraints effectively function as an information bottleneck that filters representation noise. By imposing distributional regularization on the latent spaces, the BOT configuration forces the model to prioritize encoding the most statistically relevant features while discarding irrelevant details. This filtering effect is particularly pronounced for the ESOL dataset, where solubility prediction depends primarily on specific functional groups and electronic properties rather than complete molecular structure, making effective feature selection crucial for accurate prediction.

The transition from BOT to ALL (complete DMMRL) provided further performance improvements, particularly for datasets with complex structure-property relationships such as BACE (1.4 percentage point ROC-AUC increase) and ClinTox (1.7 percentage point increase). This demonstrates that the alignment and orthogonality constraints contribute meaningfully to representation quality by enforcing statistical independence between shared and private factors while ensuring consistency across modalities. For datasets like BACE, where binding affinity depends on precise spatial arrangements and electronic complementarity with the target protein, the additional constraints help isolate the most pharmacologically relevant molecular features by aligning consistent structural patterns across modalities while separating them from modality-specific artifacts.

Notably, the complete DMMRL configuration also demonstrated substantially lower prediction variance compared to the partial configurations, particularly for the BBBP and Tox21 datasets. This variance reduction indicates that the full set of constraints not only improves average performance but also enhances model robustness and reliability—a critical consideration for applications in drug discovery where consistent predictions across diverse chemical scaffolds are essential for effective compound prioritization and development decision-making.

\subsection{\textbf{Discussion}}
Our experimental results clearly demonstrate that even without employing more sophisticated feature encoders, the feature disentanglement strategy alone yields substantial performance improvements. Specifically, while SGGRL employs contrastive learning to align multi-modal features after feature extraction(i.e., sequence, graph, and geometry encoders), our work utilizes the same multi-modal encoders as SGGRL but incorporates a shared-private feature disentanglement strategy, enabling more precise extraction of property-relevant features. This is further confirmed by our ablation experiments, which systematically validate the effectiveness of the disentanglement approach.

DMMRL achieves the best performance on five out of seven benchmark datasets, clearly outperforming existing methods in the majority of property prediction tasks. The consistent performance improvements across diverse datasets strongly suggest that molecular features are inherently entangled, and conventional alignment and fusion strategies are insufficient compared to our disentanglement approach. Notably, the advantage of DMMRL is most pronounced on datasets with high structural diversity and limited training samples, indicating that disentanglement provides a particularly strong inductive bias when learning from heterogeneous or sparse data. By focusing on the most generalizable structural features in the shared space while isolating modality-specific artifacts in private spaces, DMMRL effectively filters out noise that would otherwise contaminate the prediction pathway.

In summary, our work provides compelling evidence that representation disentanglement is a highly effective strategy for advancing molecular property prediction, and may be even more impactful than simply developing increasingly complex encoder architectures. The ability to systematically separate shared property-relevant features from private modality-specific information establishes a new paradigm for interpretable and robust multi-modal molecular representation learning.

\section{\textbf{Conclusion}}\label{sec4} 
In this paper, we propose DMMRL, directly addressing the dual challenges of representation entanglement and inadequate multi-modal integration identified in molecular property prediction. By combining feature disentanglement through variational autoencoders with adaptive fusion mechanisms, our approach enhances both predictive performance and interpretability. Extensive experiments across diverse benchmark datasets demonstrate that DMMRL not only achieves state-of-the-art results but also produces more stable and generalizable predictions, especially on structurally diverse and data-sparse tasks. The ablation studies further validate the necessity of each module, showing that variational regularization, alignment, and orthogonality constraints collectively contribute to robust and interpretable molecular representations.

Future work includes incorporating domain knowledge into regularization terms to further guide the disentanglement process and improve model interpretability. Enhancing fusion mechanisms to capture higher-order and non-linear interactions among modalities may unlock deeper insights into complex structure-property relationships. Adapting the framework for multi-task learning could enable simultaneous prediction of multiple molecular properties, increasing practical utility in drug discovery and materials science. Additionally, while this work focuses on random splitting for evaluation, exploring scaffold-based splitting could further assess the model's ability to generalize to novel molecular scaffolds and unseen chemical spaces, providing a more rigorous test of real-world applicability. Overall, DMMRL lays a solid foundation for future research in interpretable and generalizable multi-modal molecular representation learning.

\bibliographystyle{IEEEtran}
\bibliography{references}

@article{transfomer,
  title={Attention is all you need},
  author={Vaswani, Ashish and Shazeer, Noam and Parmar, Niki and Uszkoreit, Jakob and Jones, Llion and Gomez, Aidan N and Kaiser, {\L}ukasz and Polosukhin, Illia},
  journal={Advances in neural information processing systems},
  volume={30},
  year={2017}
}

@article{RNN,
  title={Empirical evaluation of gated recurrent neural networks on sequence modeling},
  author={Chung, Junyoung and Gulcehre, Caglar and Cho, KyungHyun and Bengio, Yoshua},
  journal={arXiv preprint arXiv:1412.3555},
  year={2014}
}

@article{GCN,
  title={Semi-Supervised Classification with Graph Convolutional Networks},
  author={Kipf, TN},
  journal={arXiv preprint arXiv:1609.02907},
  year={2016}
}

@article{Wrave,
  title={Molecular graph convolutions: moving beyond fingerprints},
  author={Kearnes, Steven and McCloskey, Kevin and Berndl, Marc and Pande, Vijay and Riley, Patrick},
  journal={Journal of computer-aided molecular design},
  volume={30},
  number={8},
  pages={595--608},
  year={2016},
  publisher={Springer}
}

@inproceedings{MPNN,
  title={Neural message passing for quantum chemistry},
  author={Gilmer, Justin and Schoenholz, Samuel S and Riley, Patrick F and Vinyals, Oriol and Dahl, George E},
  booktitle={International conference on machine learning},
  pages={1263--1272},
  year={2017},
  organization={Pmlr}
}

@article{n-Gram,
  title={N-gram graph: Simple unsupervised representation for graphs, with applications to molecules},
  author={Liu, Shengchao and Demirel, Mehmet F and Liang, Yingyu},
  journal={Advances in neural information processing systems},
  volume={32},
  year={2019}
}

@article{GROVER,
  title={Self-supervised graph transformer on large-scale molecular data},
  author={Rong, Yu and Bian, Yatao and Xu, Tingyang and Xie, Weiyang and Wei, Ying and Huang, Wenbing and Huang, Junzhou},
  journal={Advances in neural information processing systems},
  volume={33},
  pages={12559--12571},
  year={2020}
}

@article{li2021trimnet,
  title={TrimNet: learning molecular representation from triplet messages for biomedicine},
  author={Li, Pengyong and Li, Yuquan and Hsieh, Chang-Yu and Zhang, Shengyu and Liu, Xianggen and Liu, Huanxiang and Song, Sen and Yao, Xiaojun},
  journal={Briefings in Bioinformatics},
  volume={22},
  number={4},
  year={2021},
  publisher={Oxford Academic}
}

@article{cai2022fp,
  title={FP-GNN: a versatile deep learning architecture for enhanced molecular property prediction},
  author={Cai, Hanxuan and ZMolecular property predictionhang, Huimin and Zhao, Duancheng and Wu, Jingxing and Wang, Ling},
  journal={Briefings in bioinformatics},
  volume={23},
  number={6},
  year={2022},
  publisher={Oxford Academic}
}

@inproceedings{guo2020graseq,
  title={GraSeq: graph and sequence fusion learning for molecular property prediction},
  author={Guo, Zhichun and Yu, Wenhao and Zhang, Chuxu and Jiang, Meng and Chawla, Nitesh V},
  booktitle={Proceedings of the 29th ACM international conference on information \& knowledge management},
  pages={435--443},
  year={2020}
}

@article{zhang2024mvmrl,
  title={MvMRL: a multi-view molecular representation learning method for molecular property prediction},
  author={Zhang, Ru and Lin, Yanmei and Wu, Yijia and Deng, Lei and Zhang, Hao and Liao, Mingzhi and Peng, Yuzhong},
  journal={Briefings in Bioinformatics},
  volume={25},
  number={4},
  pages={bbae298},
  year={2024},
  publisher={Oxford University Press}
}

@article{wang2024multi,
  title={Multi-modal representation learning for molecular property prediction: sequence, graph, geometry},
  author={Wang, Zeyu and Jiang, Tianyi and Wang, Jinhuan and Xuan, Qi},
  journal={arXiv preprint arXiv:2401.03369},
  year={2024}
}

@article{schneider2020rethinking,
  title={Rethinking drug design in the artificial intelligence era},
  author={Schneider, Petra and Walters, W Patrick and Plowright, Alleyn T and Sieroka, Norman and Listgarten, Jennifer and Goodnow Jr, Robert A and Fisher, Jasmin and Jansen, Johanna M and Duca, Jos{\'e} S and Rush, Thomas S and others},
  journal={Nature reviews drug discovery},
  volume={19},
  number={5},
  pages={353--364},
  year={2020},
  publisher={Nature Publishing Group UK London}
}

@article{yang2021deep,
  title={Deep molecular representation learning via fusing physical and chemical information},
  author={Yang, Shuwen and Li, Ziyao and Song, Guojie and Cai, Lingsheng},
  journal={Advances in neural information processing systems},
  volume={34},
  pages={16346--16357},
  year={2021}
}

@inproceedings{peng2019top,
  title={Top: Towards better toxicity prediction by deep molecular representation learning},
  author={Peng, Yuzhong and Zhang, Ziqiao and Jiang, Qizhi and Guan, Jihong and Zhou, Shuigeng},
  booktitle={2019 IEEE International Conference on Bioinformatics and Biomedicine (BIBM)},
  pages={318--325},
  year={2019},
  organization={IEEE}
}

@article{liao2023sc2mol,
  title={Sc2Mol: a scaffold-based two-step molecule generator with variational autoencoder and transformer},
  author={Liao, Zhirui and Xie, Lei and Mamitsuka, Hiroshi and Zhu, Shanfeng},
  journal={Bioinformatics},
  volume={39},
  number={1},
  pages={btac814},
  year={2023},
  publisher={Oxford University Press}
}

@article{hu2024survey,
  title={A survey on information bottleneck},
  author={Hu, Shizhe and Lou, Zhengzheng and Yan, Xiaoqiang and Ye, Yangdong},
  journal={IEEE Transactions on Pattern Analysis and Machine Intelligence},
  volume={46},
  number={8},
  pages={5325--5344},
  year={2024},
  publisher={IEEE}
}

@inproceedings{dey2017gate,
  title={Gate-variants of gated recurrent unit (GRU) neural networks},
  author={Dey, Rahul and Salem, Fathi M},
  booktitle={2017 IEEE 60th international midwest symposium on circuits and systems (MWSCAS)},
  pages={1597--1600},
  year={2017},
  organization={IEEE}
}

@article{walters2020applications,
  title={Applications of deep learning in molecule generation and molecular property prediction},
  author={Walters, W Patrick and Barzilay, Regina},
  journal={Accounts of chemical research},
  volume={54},
  number={2},
  pages={263--270},
  year={2020},
  publisher={ACS Publications}
}

@article{li2022deep,
  title={Deep learning methods for molecular representation and property prediction},
  author={Li, Zhen and Jiang, Mingjian and Wang, Shuang and Zhang, Shugang},
  journal={Drug Discovery Today},
  volume={27},
  number={12},
  pages={103373},
  year={2022},
  publisher={Elsevier}
}

@article{xia2023understanding,
  title={Understanding the limitations of deep models for molecular property prediction: Insights and solutions},
  author={Xia, Jun and Zhang, Lecheng and Zhu, Xiao and Liu, Yue and Gao, Zhangyang and Hu, Bozhen and Tan, Cheng and Zheng, Jiangbin and Li, Siyuan and Li, Stan Z},
  journal={Advances in Neural Information Processing Systems},
  volume={36},
  pages={64774--64792},
  year={2023}
}

@article{wu2023molecular,
  title={Molecular joint representation learning via multi-modal information of SMILES and graphs},
  author={Wu, Tianyu and Tang, Yang and Sun, Qiyu and Xiong, Luolin},
  journal={IEEE/ACM transactions on computational biology and bioinformatics},
  volume={20},
  number={5},
  pages={3044--3055},
  year={2023},
  publisher={IEEE}
}

@article{wu2023improved,
  title={An improved multi-modal representation-learning model based on fusion networks for property prediction in drug discovery},
  author={Wu, Jinzhou and Su, Yang and Yang, Ao and Ren, Jingzheng and Xiang, Yi},
  journal={Computers in biology and medicine},
  volume={165},
  pages={107452},
  year={2023},
  publisher={Elsevier}
}

@inproceedings{xu2021multi,
  title={Multi-VAE: Learning disentangled view-common and view-peculiar visual representations for multi-view clustering},
  author={Xu, Jie and Ren, Yazhou and Tang, Huayi and Pu, Xiaorong and Zhu, Xiaofeng and Zeng, Ming and He, Lifang},
  booktitle={Proceedings of the IEEE/CVF international conference on computer vision},
  pages={9234--9243},
  year={2021}
}

@inproceedings{lee2021private,
  title={Private-shared disentangled multimodal vae for learning of latent representations},
  author={Lee, Mihee and Pavlovic, Vladimir},
  booktitle={Proceedings of the ieee/cvf conference on computer vision and pattern recognition},
  pages={1692--1700},
  year={2021}
}

@article{niu2021review,
  title={A review on the attention mechanism of deep learning},
  author={Niu, Zhaoyang and Zhong, Guoqiang and Yu, Hui},
  journal={Neurocomputing},
  volume={452},
  pages={48--62},
  year={2021},
  publisher={Elsevier}
}

@article{weerakody2021review,
  title={A review of irregular time series data handling with gated recurrent neural networks},
  author={Weerakody, Philip B and Wong, Kok Wai and Wang, Guanjin and Ela, Wendell},
  journal={Neurocomputing},
  volume={441},
  pages={161--178},
  year={2021},
  publisher={Elsevier}
}

@misc{wu2018moleculenetbenchmarkmolecularmachine,
      title={MoleculeNet: A Benchmark for Molecular Machine Learning}, 
      author={Zhenqin Wu and Bharath Ramsundar and Evan N. Feinberg and Joseph Gomes and Caleb Geniesse and Aneesh S. Pappu and Karl Leswing and Vijay Pande},
      year={2018},
      eprint={1703.00564},
      archivePrefix={arXiv},
      primaryClass={cs.LG},
}

@misc{ba2016layernormalization,
      title={Layer Normalization}, 
      author={Jimmy Lei Ba and Jamie Ryan Kiros and Geoffrey E. Hinton},
      year={2016},
      eprint={1607.06450},
      archivePrefix={arXiv},
      primaryClass={stat.ML},
      url={https://arxiv.org/abs/1607.06450}, 
}

@inproceedings{chen2020dynamic,
  title={Dynamic relu},
  author={Chen, Yinpeng and Dai, Xiyang and Liu, Mengchen and Chen, Dongdong and Yuan, Lu and Liu, Zicheng},
  booktitle={European conference on computer vision},
  pages={351--367},
  year={2020},
  organization={Springer}
}

@article{kingma2014adam,
  title={Adam: A method for stochastic optimization},
  author={Kingma, Diederik P},
  journal={arXiv preprint arXiv:1412.6980},
  year={2014}
}

@inproceedings{he2016deep,
  title={Deep residual learning for image recognition},
  author={He, Kaiming and Zhang, Xiangyu and Ren, Shaoqing and Sun, Jian},
  booktitle={Proceedings of the IEEE conference on computer vision and pattern recognition},
  pages={770--778},
  year={2016}
}

\end{document}